# LLMCARE: early detection of cognitive impairment via transformer models enhanced by LLM-generated synthetic data

**Ali Zolnour, MS[1]; Hossein Azadmaleki, MS[1]; Yasaman Haghbin, MS[1]; Fatemeh Taherinezhad, MS[1]; Mohamad Javad Momeni Nezhad, MS[1]; Sina Rashidi, MS[1]; Masoud Khani, MS[2]; AmirSajjad Taleban, MS[2]; Samin Mahdizadeh Sani, MS[3]; Maryam Dadkhah, MS[1]; James M. Noble, MD[1,4]; Suzanne Bakken, PhD[5,6,7]; Yadollah Yaghoobzadeh, PhD[3]; Abdol-Hossein Vahabie, PhD[3]; Masoud Rouhizadeh[8], PhD; Maryam Zolnoori, PhD[1,5,7]**

[1]Columbia University Irving Medical Center, New York, NY, United States
[2]University of Wisconsin-Milwaukee, Milwaukee, WI, United States
[3]School of Electrical and Computer Engineering, University of Tehran, Tehran, IRAN
[4]Department of Neurology, Taub Institute for Research on Alzheimer's Disease and the Aging Brain, GH Sergievsky Center, Columbia University, New York, NY, United States
[5]School of Nursing, Columbia University, New York, NY 10032, United States
[6]Department of Biomedical Informatics, Columbia University, New York, NY 10032, United States
[7]Data Science Institute, Columbia University, New York, NY 10027, United States
[8]University of Florida, College of Pharmacy, Gainesville, FL, United States

**Corresponding Author:**

Maryam Zolnoori, PhD

Columbia University Medical Center,

School of Nursing Columbia University

560 W 168th St, New York, NY 10032

Phone: 317-515-1950

E-mail: mz2825@cumc.columbia.edu; m.zolnoori@gmail.com

**Number of Tables:** 4

**Number of Figures:** 6

**Conflict of Interest:** Authors have no conflict of interest to declare.

**Clinical Trial Number:** Not applicable.

**Funding Declaration:**

- R00AG076808: "Development of a Screening Algorithm for Timely Identification of Patients with Mild Cognitive Impairment and Early Dementia in Home Healthcare" from National Institute on Aging.
- P30AG059303: Columbia Center for Interdisciplinary Research on Alzheimer's Disease Disparities (CIRAD)

**Code Availability:** The codes for LLMCARE are publicly available at: GitHub

**Data Availability statement:** The data are available in two sources: (i) the ADReSSo 2021 benchmark dataset, derived from the Pitt Corpus in DementiaBank, comprising 237 participants labeled as cognitively impaired or cognitively healthy, and (ii) the Delaware corpus, also derived from DementiaBank, comprising 205 English-speaking participants labeled as mild cognitive impairment (MCI) or cognitively healthy.

## Abstract

**Background:** Alzheimer's disease and related dementias (ADRD) affect nearly five million older adults in the United States, yet more than half remain undiagnosed. Speech-based natural language processing (NLP) provides a scalable approach to identify early cognitive decline by detecting subtle linguistic markers that may precede clinical diagnosis.

**Objective:** This study aims to develop and evaluate a speech-based screening pipeline that integrates transformer-based embeddings with handcrafted linguistic features, incorporates synthetic augmentation using large language models (LLMs), and benchmarks unimodal and multimodal LLM classifiers. External validation was performed to assess generalizability to an MCI-only cohort.

**Methods:** Transcripts were obtained from the ADReSSo 2021 benchmark dataset (n = 237; derived from the Pitt Corpus, DementiaBank) and the DementiaBank Delaware corpus (n = 205; clinically diagnosed mild cognitive impairment [MCI] vs controls). Audio was automatically transcribed using Amazon Web Services Transcribe (general model). Ten transformer models were evaluated under three fine-tuning strategies. A late-fusion model combined embeddings from the best-performing transformer with 110 linguistically derived features. Five LLMs (LLaMA-8B/70B, MedAlpaca-7B, Ministral-8B, GPT-4o) were fine-tuned to generate label-conditioned synthetic speech for data augmentation. Three multimodal LLMs (GPT-4o, Qwen-Omni, Phi-4) were tested in zero-shot and fine-tuned settings.

**Results:** On the ADReSSo dataset, the fusion model achieved an F1-score of 83.32 (AUC = 89.48), outperforming both transformer-only and linguistic-only baselines. Augmentation with MedAlpaca-7B synthetic speech improved performance to F1 = 85.65 at 2× scale, whereas higher augmentation volumes reduced gains. Fine-tuning improved unimodal LLM classifiers (e.g., MedAlpaca-7B, F1 = 47.73 → 78.69), while multimodal models demonstrated lower performance (Phi-4 = 71.59; GPT-4o omni = 67.57). On the Delaware corpus, the pipeline generalized to an MCI-only cohort, with the fusion model plus 1× MedAlpaca-7B augmentation achieving F1 = 72.82 (AUC = 69.57).

**Conclusions:** Integrating transformer embeddings with handcrafted linguistic features enhances ADRD detection from speech. Distributionally aligned LLM-generated narratives provide effective but bounded augmentation, while current multimodal models remain limited. Crucially, validation on the Delaware corpus demonstrates that the proposed pipeline generalizes to early-stage impairment, supporting its potential as a scalable approach for clinically relevant early screening. All codes for LLMCARE are publicly available at: GitHub

**Keyword:** Alzheimer's disease, mild cognitive impairment (MCI), large language models, data augmentation, transformers, natural language processing

## 1. INTRODUCTION

Alzheimer's disease and related dementias (ADRD) pose a major public health challenge, affecting approximately five million individuals—11% of older adults—in the United States[1,2]. Despite national efforts, over half of patients remain undiagnosed and untreated[3–5]. With an expected 13.2 million cases by 2050[6], the National Institute on Aging has prioritized the development of effective screening tools[7,5]. Meeting this need requires an interdisciplinary approach spanning neuroscience, data science, and speech-language pathology.

One promising direction involves leveraging natural language processing (NLP) to analyze spontaneous speech,[8] which can reveal subtle cognitive changes often missed by traditional screening instruments. Early linguistic impairments—such as word-finding difficulties[9,10], syntactic disorganization[11], and reduced fluency[9]—may be detectable through tasks like picture descriptions. Although speech-based screening has shown potential, progress is limited by scarce labeled clinical speech data and poor model generalizability across populations and clinical settings.[12,13]

Transformer-based NLP models—particularly BERT[14] and its variants—capture linguistic context well and have achieved strong results in classifying cognitive impairment in corpora such as DementiaBank[15]. However, variations in fine-tuning protocols, validation sets, and downstream classifiers lead to inconsistent findings on how well these models encode linguistic markers of cognitive decline (see Table 4 in Appendix A as an example of this variation).[2] Progress is further constrained by the small size of available speech datasets, which limits both model training and rigorous model validation.

Recent work suggests that large language models (LLMs), such as GPT-4[16], can generate synthetic clinical data resembling real-world datasets. Compared to generative adversarial networks[17], LLMs are more accessible and require less technical expertise. Yet, their effectiveness for downstream tasks varies. For instance, synthetic mental health interviews significantly improved ML-based depression detection[18], while only marginal gains were observed for named entity recognition in social determinants of health (e.g., Macro-F1 improvement <1%)[19]. In autism detection, synthetic data increased recall by 13% but reduced precision by 16%[20]. These mixed results highlight that LLM-generated data must preserve linguistic complexity, align with real data distributions, and support generalization.

Beyond text, emerging multimodal LLMs extend these capabilities by jointly modeling language and audio inputs, enabling them to capture both what is said and how it is said—such as prosody[21]. These acoustic-linguistic features may be critical in early cognitive impairment detection. However, their application in dementia research remains limited.

This study addresses these gaps through a multi-component design evaluated on two datasets: the ADReSSo 2021 benchmark, which includes participants across a range of cognitive impairment severity (from mild cognitive impairment [MCI] to severe dementia) versus cognitively healthy [controls], and the Delaware corpus, which is restricted to clinically diagnosed MCI versus controls.

Component 1. Developing the Screening Algorithm: We systematically evaluated BERT-based and newer transformers (e.g., BGE) on the picture-description task to identify the optimal model for encoding linguistic cues. We then combined embeddings from the top-performing model with handcrafted features (e.g., lexical richness) to develop a screening algorithm, hypothesizing that integration would enhance detection accuracy.

Component 2. Leveraging LLMs to Generate Synthetic Speech: We evaluated state-of-the-art LLMs, including open-weight (LLaMA, MedAlpaca, Ministral) and commercial (GPT-4), to assess their ability to learn linguistic markers of cognitive impairment and generate synthetic speech faithful to patient language. We then tested whether augmenting training data with synthetic speech improved screening performance.

Component 3. Evaluation of LLMs as Classifiers: We assessed the diagnostic capabilities of LLMs in zero-shot and fine-tuned settings to establish baseline and advanced benchmarks, examining whether model size and training improve classification compared to pre-trained transformers.

Component 4. Evaluating Multimodal LLMs for Integrated Speech and Text Analysis: We explored whether multimodal LLMs that jointly process linguistic and acoustic inputs improve detection of cognitive impairment compared to text-only models.

Component 5. Validation on the Delaware Corpus: Finally, we validated the pipeline on the Delaware dataset, restricted to clinically diagnosed MCI versus controls, to assess generalizability beyond the mixed-severity cohort in ADReSSo 2021 and to test the performance of Components 1–4 in early-stage cognitive impairment screening.

This study makes several contributions to speech-based ADRD detection. We systematically evaluate ten transformer architectures on the ADReSSo 2021 benchmark and show that combining transformer embeddings with 110 linguistic features in a late-fusion design improves generalization. We introduce a controlled augmentation framework using distributionally aligned LLM-generated narratives, which enhances performance without compromising validity. We also benchmark unimodal and multimodal LLMs, demonstrating that well-tuned linguistic transformers remain competitive with large-scale LLMs. Finally, we validate the pipeline on the Delaware dataset, the first evaluation of this approach on clinically diagnosed MCI versus controls, providing evidence of generalizability to early-stage impairment.

## 2. METHOD

### 2.1 Pipeline Overview

Figure 1 illustrates the methodology for developing the screening algorithm using the fusion of pre-trained transformer model and handcrafted lexical features, process of synthetic text generation using state-of-the-art LLMs, and measuring the performance of both unimodal and multimodal LLMs as classifiers for ADRD detection.

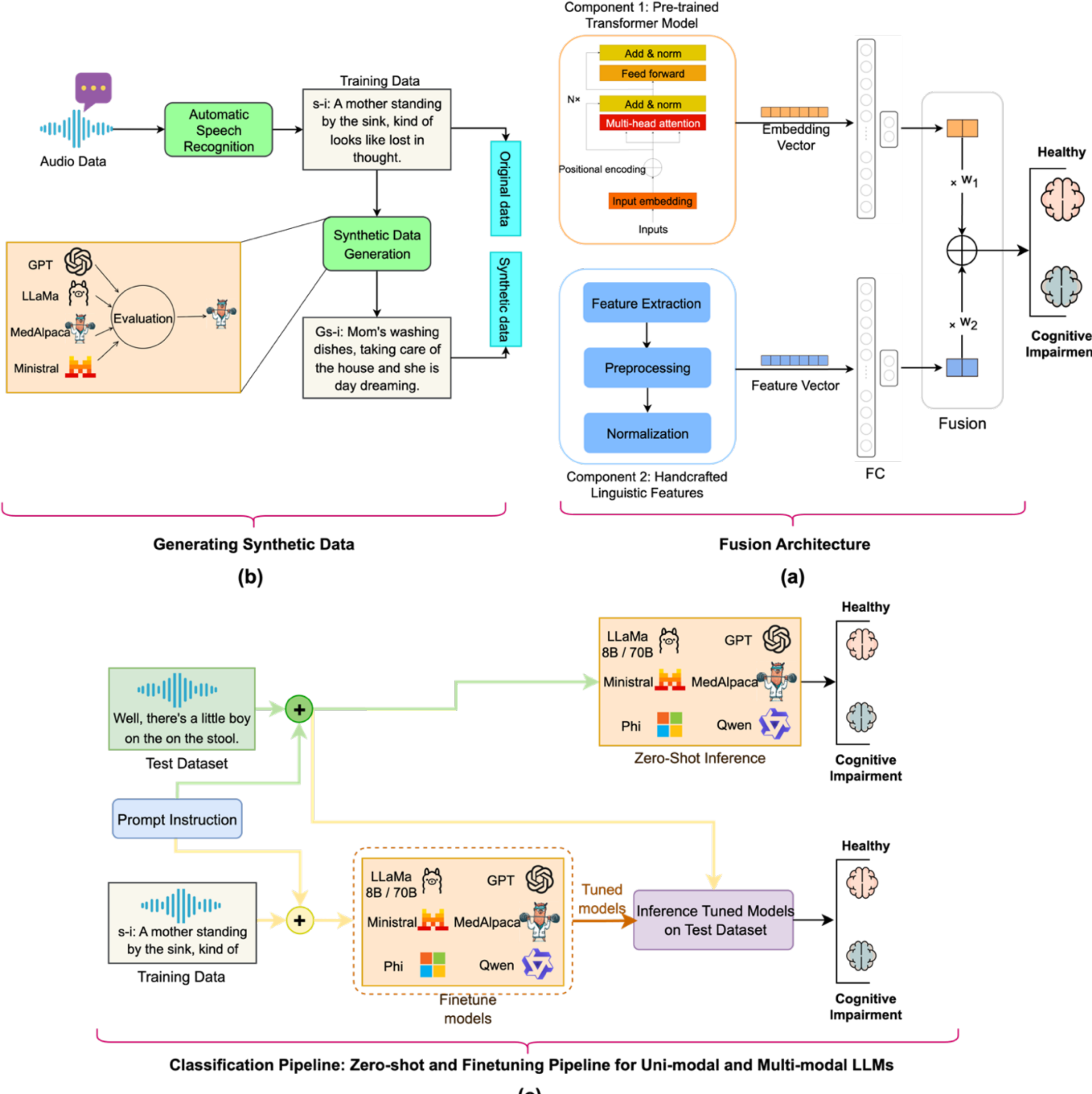

**Figure 1.** Overview of the study's methodology: (a) Fusion model: Automatic speech-recognition transcripts are transformed into deep transformer embeddings and merged with handcrafted linguistic features in a fully connected fusion layer; (b) Generative synthetic data: GPT-4, LLaMA-8B/70B, MedAlpaca-7B, and Ministral-8B create additional transcripts that mirror the original distribution for augmenting the training dataset; (c) Classification pipeline – Zero-shot and fine-tuning for unimodal and multimodal LLMs: Both unimodal and multimodal LLMs were evaluated in both zero-shot and fine-tuning regimes to assign "cognitively healthy" or "cognitively impaired" labels on a held-out test set.

### 2.2 Dataset and Cohorts

We used the ADReSSo 2021 benchmark dataset, derived from DementiaBank's Pitt Corpus and introduced as a standardized benchmark for dementia detection. The dataset includes 237 participants performing the Cookie-Theft picture description and is divided into an official training set (166 participants) and an official test set (71 participants), balanced for age and gender. For model development, we further split the training portion into 116 training and 50 validation participants using stratified sampling, with all hyperparameter tuning performed on the validation set only. The final results are reported on the official ADReSSo 2021 test set, ensuring comparability with prior work.

Participant characteristics are summarized in Table 1. All individuals underwent comprehensive neuropsychological evaluations, including verbal tasks and the Mini-Mental State Examination (MMSE). Diagnoses were assigned by clinical specialists (neurologists and neuropsychologists) following full clinical assessments. All participants were older than 53 years, and females comprised more than 60% of each group. MMSE scores in the case group ranged from 7–28 (training), 3–27 (validation), and 5–27 (test), reflecting mild to severe cognitive impairment, while scores in the control group ranged from 24–30, consistent with normal cognition. On average, control participants produced more words and had shorter recordings than cases, consistent with expected language production differences in dementia.

**Table 1**. Baseline demographic, cognitive, and speech characteristics of participants across training, validation, and test cohorts. Recording length is reported in seconds (s).

| Attribute | Train | | Validation | | Test | |
|---|---|---|---|---|---|---|
| | Case | Control | Case | Control | Case | Control |
| Participants (N) | 60 | 56 | 27 | 23 | 35 | 36 |
| Gender (F/M) | 39/21 | 37/19 | 19/8 | 15/8 | 21/14 | 23/13 |
| Age (Mean ± Std) | 69.33 ± 7.14 | 66.27 ± 6.81 | 70.59 ± 6.01 | 65.48 ± 4.72 | 68.51 ± 7.12 | 66.11 ± 6.53 |
| Age Range | 53–79 | 54–80 | 60–80 | 56–74 | 56–79 | 56–78 |
| MMSE (Mean ± Std) | 17.80 ± 5.04 | 29.04 ± 1.13 | 16.63 ± 5.94 | 28.87 ± 1.22 | 18.86 ± 5.8 | 28.91 ± 1.25 |
| MMSE Range | 7–28 | 26–30 | 3–27 | 26–30 | 5–27 | 24–30 |
| Recording Length (s), (Mean ± Std) | 87.20 ± 48.35 | 68.98 ± 25.85 | 88.52 ± 43.27 | 68.25 ± 25.43 | 79.42 ± 36.79 | 66.35 ± 28.17 |

| Recording Length (s), Range | 35.26–268.49 | 22.79–168.61 | 39.91–219.5 | 26.16–121.47 | 28.39–150.15 | 22.35–135.68 |
|---|---|---|---|---|---|---|
| Word Count (Mean ± Std) | 82 ± 43 | 114 ± 78 | 101 ± 55 | 111 ± 43 | 92 ± 57 | 111 ± 53 |
| Word Count Range | 22–189 | 21–523 | 31–284 | 54–197 | 27–256 | 45–243 |

**Transcription and Preprocessing.** To avoid reliance on manual transcripts, we re-transcribed the audio data with Amazon Transcribe (general model). We minimized normalization (automatic grammar rewriting disabled), enabled disfluency/hesitation tokens and partial-word emission, and retained word-level timestamps. All study transcripts were produced with this configuration without human edits.

In prior benchmarking on the same audio against Amazon Transcribe (medical), Whisper-Large, and a fine-tuned wav2vec2, the Amazon general model achieved a competitive English Word Error Rate (WER) ≈ 13% and most faithfully preserved verbatim phenomena—fillers (“uh/um”), lexical repetitions, and fragmented/partial words—compared with the alternatives. These cues are central to our feature extraction and fusion classifier.

### 2.3 Component 1. Developing the Screening Algorithm Using Pre-Trained Transformer Models and Domain-Related Linguistic Features

#### Linguistic Transformers Baselines

We systematically evaluated ten transformer models commonly cited in healthcare NLP literature to assess their ability to detect subtle linguistic cues indicative of cognitive decline. Leveraging attention mechanisms, transformers can identify disfluencies such as repetitions, syntactic errors, and filler words—key linguistic cues of cognitive impairment.

Our evaluation included five general-purpose models—BERT, DistilBERT [22], RoBERTa [23], XLNet [24], BGE[25], and Longformer [26] —pretrained on corpora like Wikipedia, BookCorpus, and online health forums. We also tested five domain-specific models—BioBERT[27], BioClinicalBERT[28], ClinicalBigBird[29], and BlueBERT[30]—trained on biomedical and clinical texts. We hypothesized that domain-specific models may be less sensitive to disfluent, conversational speech, limiting their ability to capture nuanced impairments.

To evaluate fine-tuning strategies, we tested three configurations: (i) no fine-tuning (frozen transformer as feature extractor), (ii) full fine-tuning (updating parameters of all layers), and (iii) last-layer fine-tuning (updating only the final transformer layer). The frozen model served as a baseline. Although full fine-tuning can boost performance, it risks overfitting on small datasets. Last-layer tuning retains ~90% of the performance gain while preserving generalizable features and reducing computational cost. Fine-tuning intermediate layers (e.g., layers 6–7 in BERT) was not pursued due to minimal added benefit and increased complexity.

Each transformer was paired with a two-layer multilayer perceptron (MLP) classifier. Embeddings were fed into the MLP with 256 hidden units and a 0.4 dropout rate. For both full fine-tuning and last-layer fine-tuning approaches, models were trained using AdamW (batch size = 8, learning rate = $2 \times 10^{-5}$, weight decay = $2 \times 10^{-3}$) for 50 epochs. We selected the best-performing epoch based on the highest F1-score on the validation dataset. To reduce variance from random initialization, each experiment was repeated five times with different seeds; Next, we reported the average F1-score on the held-out test set using the best validation epoch.

**Handcrafted Linguistic Features**

Transformer embeddings can detect linguistic patterns but often lack transparency. To improve interpretability, we extracted 110 handcrafted lexical features across four dimensions[2]: (1) lexical richness was measured with established diversity metrics to gauge reliance on high-frequency vocabulary; [31,32,33] (2) syntactic complexity was assessed through part-of-speech tagging to reflect grammatical structure; [34,35] (3) semantic coherence and fluency were quantified by measuring word repetition and filler words; [36,37] (4) psycholinguistic cues were extracted using LIWC 2015, which groups commonly used words into 11 top-level categories (e.g., affective, social, cognition) relevant to cognitive decline (see Appendix B for details of these lexical features). [38–42]

We built a lexical feature-based model using 110 lexical features as input and two-layer MLP with 64 hidden neurons, trained with AdamW (learning rate = $8 \times 10^{-3}$, weight decay = $1 \times 10^{-3}$) for up to 50 epochs. We selected the best-performing epoch based on the highest F1-score on the validation dataset. This allowed us to evaluate the standalone utility of handcrafted features.

**Late Fusion Classifier**

To combine the strengths of transformer representations and domain-informed features, we developed a fusion classifier (Figure 1). Embeddings from the top-performing transformer were passed through a two-layer MLP with 256 hidden units; linguistic features entered a separate two-layer MLP with 128 units. We applied a late fusion strategy by combining the two outputs using a learnable weighted sum. The fusion model was trained using AdamW (learning rate = $2 \times 10^{-5}$, weight decay = $2 \times 10^{-3}$) for 50 epochs. Each experiment was repeated five times with different random seeds. We report the mean and 95% confidence intervals of F1-score and AUC-ROC on the validation and test sets.

## 2.4. Component 2. LLM-Based Synthetic Text for Augmentation

To generate synthetic descriptions reflecting speech of cognitively impaired or cognitively healthy, we adopted a label-conditioned language modeling framework, where each token is generated based on the prior context and the target label. This approach allows LLMs to learn and reproduce label-specific linguistic features—such as repetition or disfluency— that are critical for data augmentation in classification tasks.

$$C = \{(S_i, y_i) | i = 1, 2, \dots, M\},$$

where each sequence $S_i = (w_1^i, w_2^i, \dots, w_{T_i}^i)$ is paired with a cognitive status label $y_i \epsilon \{Case,\ Control\}$, our goal is to model the conditional probability

$$P_\theta(S_i | y_i) = \prod_{t=1}^{T_i} P_\theta(w_t^i | w_1^i, \dots, w_{t-1}^i, y_i)$$

It is important to note that synthetic transcriptions were generated exclusively for training augmentation; all validation and test evaluations were conducted on real, held-out participants.

**LLM Models and Tuning**

We evaluated five LLMs spanning model sizes and training data: LLaMA 3.1 8B Instruct[43]: Balanced in size and quality, suitable for generating coherent narratives with class-specific variation (e.g., reduced vocabulary); MedAlpaca 7B[44]: A clinically fine-tuned model included to test whether exposure to biomedical language improves generation of patient-like language and terminology alignment; Ministral 8B Instruct[45]: Offers strong sentence-level coherence and low-latency inference, suitable for generating fluent but compact narratives typical of non-cognitively impaired individuals; LLaMA 3.3 70B Instruct: Used to evaluate whether increased model capacity improves simulation of complex or disorganized language patterns in cognitively impaired speech; GPT-4o[46] (text-only mode): Used as a benchmark for fluency and coherence, capable of mimicking subtle disfluencies when prompted.

We fine-tuned open-weight LLMs (LLaMA 3.1 8B, MedAlpaca 7B, Ministral 8B, and LLaMA 3.3 70B) using the Quantized Low-Rank Adapter (QLoRA) framework, which inserts lightweight adapters into frozen models to enable memory-efficient training. We tested LoRA ranks of 64 and 128, with scaling factors set to α = 2 × rank, and applied dropout rates of 0 or 0.1 within adapter layers to mitigate overfitting. For LLaMA 3.1 8B, MedAlpaca 7B, and Ministral 8B, adapters were inserted into all linear layers. For LLaMA 3.3 70B, model weights were quantized to 4-bit precision before fine-tuning, and adapters were placed only in the query, key, and value (QKV) projection layers to reduce memory usage. Other fine-tuning parameters for open-weight LLMs included the PagedAdamW optimizer with mixed-precision (float16) training, a cosine learning rate scheduler, and learning rates of 2e-4 or 1e-4. For GPT-4o, only batch size (16 or 20) and learning rate multipliers (2.5 or 3) were tuned. All LLMs were trained for 10 epochs.

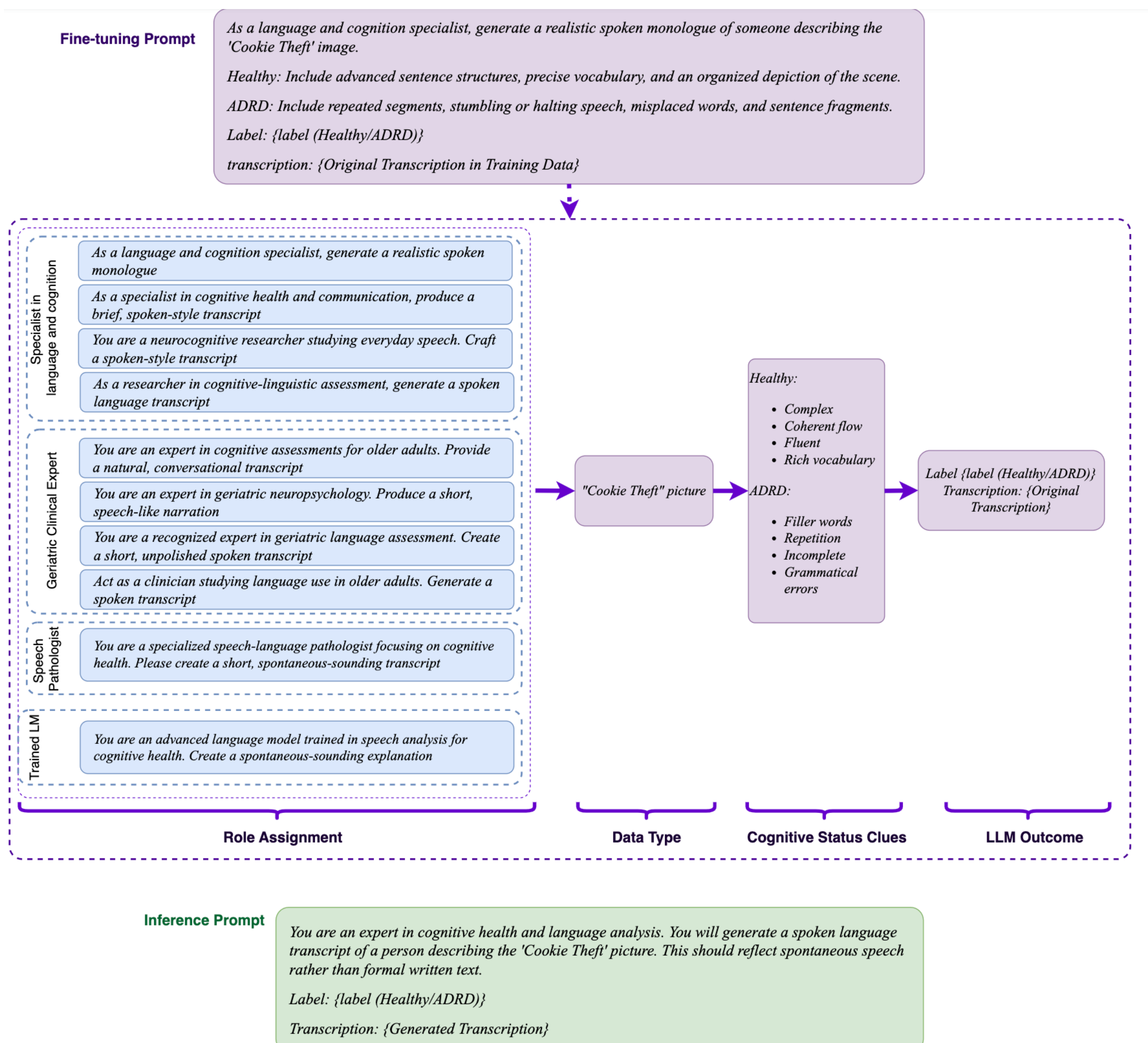

**Figure 2.** Prompt-engineering workflow for synthetic transcript generation and classification. Fine-tuning prompt: A role-specific instruction directs the LLM to describe the Cookie-Theft picture in spoken language, embedding class-defining cues—advanced syntax and fluent flow for cognitively healthy [Healthy] speech, repetition and grammatical slips for cognitively impaired [ADRD] speech; Role templates: Ten expert personas (e.g., language-and-cognition specialist, geriatric clinician, speech-language pathologist) provide prompt diversity while the task wording remains constant; Data & cues: The prompt explicitly references the Cookie-Theft image and the cognitive cues the model should express; Output: Each generated transcript is saved with its label (Healthy or ADRD);

Inference prompt: A neutral expert persona requests a transcript label without repeating class hints, encouraging the model to rely on patterns learned during fine-tuning.

## Prompt Design and Inference

For fine-tuning, we used prompts that incorporated label-specific linguistic cues—for example, “advanced sentence structures” for cognitively healthy (control) participants and “repetition and filler words” for cognitively impaired (case) participants. This improved the model’s ability to generate class-consistent outputs. However, relying on a single fixed prompt reduced transcription diversity and limited generalizability. To address this, we created 10 prompt variations that differed in the assigned role (e.g., “language and cognition specialist” vs. “speech pathologist”) while keeping task instructions consistent. With 116 training samples, each prompt was applied to about 11–12 samples, ensuring controlled variation without introducing excessive noise (see Figure 2).

For inference, we initially tested prompts that also included label-specific cues, but these produced repetitive and unnatural outputs. To encourage more spontaneous and generalizable speech, we instead used neutral prompts, allowing models to rely on the linguistic patterns learned during fine-tuning (see Figure 2; also for more details about prompt engineering see Appendix C). Inference hyperparameters were tuned to balance coherence and diversity of generated text. We tested values for top-p (0, 0.9, 0.95, –1), top-k (40, 50), and temperature (0.5, 0.7, 0.9, 1.0, 2.0). Optimal settings were: top-p = 0.95, top-k = 50, and temperature = 1 for LLaMA-8B, LLaMA-70B, and MedAlpaca-7B; top-p disabled for Ministral-8B; and temperature = 1 for GPT-4o, consistent with OpenAI’s single-parameter guidance.

## Synthetic Data Evaluation and Scaling

Evaluation metrics for measuring the quality of the synthetic generated data included:

1. **F1-score on Validation Dataset:** For each LLM and fine-tuning epoch, we generated a synthetic dataset (N = 116) using the inference prompt. We then retrained the fusion-based screening algorithm on the combined original training data and synthetic data and measured its performance on the validation set. The highest F1-score determined the optimal configuration for each LLM. The optimal configuration for each LLM is presented in Table 5 in Appendix B**.**

2. **BLEU[47] and BERTScore[48]:** BLEU measured syntactic similarity by computing n-gram overlap (n = 1–4) between generated and reference transcriptions in the validation dataset, while BERTScore assessed semantic similarity using contextualized embeddings. These metrics provided additional insight into the extent to which the generated transcriptions preserved structural and semantic properties of original patient speech.

3. **t-SNE**[49] **Visualization:** We applied t-SNE to sentence-level embeddings from synthetic data, original training set, the validation set, and the held-out test set to visualize overlap and distribution similarity in embedding space.

Using the best configuration for each LLM, we generated synthetic data at 1x to 5x the size of the original training set and measured the fusion-based screening model's performance. This assessed whether larger volumes of synthetic text enhanced generalizability while preserving diagnostic cues

### 2.5 Component 3. LLMs as Classifiers (Text-Only)

We evaluated whether LLMs—LLaMA (variants), MedAlpaca, Ministral, and GPT-4—can classify transcripts as "Cognitively healthy" or "Cognitively impaired" both without task-specific training (zero-shot) and with fine-tuning.

**Zero-shot prompting**. To identify an effective prompt, we tested several prompting formulations and selected one that: (a) assigns the model the role of a cognitive-and-language expert; (b) specifies that the input is a transcript of spontaneous speech; (c) instructs a binary decision ("cognitively healthy" vs "cognitively impaired"); and (d) omits explicit linguistic cues, encouraging the model to rely on its internal reasoning and general language knowledge (see Appendix B for the exact prompt). For inference, open-weight models used temperature = 0 (deterministic outputs), and GPT-4 used temperature = 0.7 per platform guidance.

**Fine-tuning**. We also fine-tuned each LLM to classify transcripts as Healthy or ADRD. Unlike Component 2, where generation and inference prompts differed, here we used the same prompt during fine-tuning and inference to promote stability and consistency. The hyperparameter search mirrored Component 2. Each model was trained for 10 epochs, and the best checkpoint was chosen by the highest validation F1-score. Final performance was reported on the held-out test set using F1-score, precision, and recall.

### 2.6 Component 4. Evaluating Multimodal LLMs as Classifiers

We evaluated three state-of-the-art audio–text multimodal models:

1. Qwen 2.5-Omni[50] (7B–8.4B parameters): an open-weight "Thinker-Talker" architecture that natively processes text, audio, image, and video, supporting real-time speech responses and full fine-tuning via Hugging Face checkpoints.
2. Phi-4-Multimodal[51] (5.6B parameters): Microsoft's successor to the Phi-series, unifying speech, vision, and language encoders into a single network, offering 128K-token context. We used its open-weight version for domain-specific fine-tuning.
3. GPT-4o ("omni"): OpenAI's flagship closed-weight model with sub-300 ms speech latency, capable of processing any mix of text, audio, image, and video. We tested it only in zero-shot mode due to unavailable API for fine-tuning.

We evaluated Qwen and Phi in both zero-shot and fine-tuning settings, whereas GPT-4o was assessed in zero-shot only. This design allowed comparison of joint acoustic-linguistic modeling against text-only baselines and evaluation of the benefits of multimodal fine-tuning.

### 2.7 Component 5. External Generalizability Evaluation — DementiaBank Delaware Corpus

We evaluated performance of the pipeline on the DementiaBank Delaware corpus, which includes three picture-description tasks (Cookie Theft, Cat Rescue, Rockwell), a Cinderella story recall, and a procedural discourse task from 205 English-speaking participants (99 MCI, 106 controls). Labels were binary (clinically diagnosed MCI vs. control).

We applied a participant-level split (~60% train: n=124; ~20% validation: n=40; ~20% test: n=41), ensuring recordings from each individual appeared in only one partition. Initial experiments showed that Cat Rescue and Rockwell tasks provided limited discriminatory signals (based on F1 scores on the validation set). We therefore focused on Cookie Theft, Cinderella recall, and procedural discourse, which yielded stronger MCI-detection performance.

#### Screening Algorithm with Transformers and Linguistic Features

(1) Pre-trained Transformer Baselines: We fine-tuned four top-performing transformers on the ADReSSo dataset—BERT, DistilBERT, Longformer, and BioBERT. Each transformer fed embeddings into a two-layer MLP classifier (256 hidden units; dropout=0.4). Last layer of models was trained with AdamW (batch size=8, learning rate=$2\times10^{-5}$, weight decay=$2\times10^{-3}$) for 50 epochs. The best epoch was selected by the highest F1-score on the validation set.

(2) Fusion of Transformer Embeddings and Handcrafted Linguistic Features: Embeddings from the top-performing transformer were passed through a two-layer MLP (256 hidden units). Handcrafted linguistic features were processed by a separate two-layer MLP (64 hidden units). We applied late fusion via a learnable weighted sum of the two outputs. The fusion model was trained with AdamW (learning rate=$2\times10^{-5}$, weight decay=$2\times10^{-3}$) for 50 epochs.

#### LLM-Based Synthetic Augmentation

(1) Leveraging LLMs to Generate Synthetic Data: We fine-tuned MedAlpaca-7B—the best model identified in ADReSSo—for data augmentation, using the same prompting strategy as for the ADReSSo dataset. MedAlpaca-7B was fine-tuned separately for each task, and synthetic samples were generated per task.

(2) Assessing Augmentation Effects: We fine-tuned DistilBERT + linguistic features with 1× and 2× augmentation to evaluate the impact of LLM-generated synthetic data on MCI detection.

#### Text-Based LLMs (Unimodal) and Multimodal (Audio +Text) LLMs Classifier

**Text-Based LLMs.** Since LLaMA 3.1 8B, LLaMA 3.3 70B, and GPT-4o emerged as the top three text-only LLMs on the ADReSSo 2021 dataset—achieving the best classification performance across both zero-shot and fine-tuning strategies—we selected these models for external evaluation on the Delaware dataset.

**Multimodal LLMs.** For multimodal classification, we opted for GPT-4o (omni) and Phi-4, again applying both zero-shot and fine-tuning strategies, as these models demonstrated strong performance and robust handling of multimodal inputs.

## 3. RESULT

### 3.1 Component 1. Developing the Screening Algorithm Using Pre-Trained Transformer Models and Domain-Related Linguistic Features

#### Performance of Transformer Models

Figure 3.A reports results for six general-purpose transformers and Figure 3.B for four clinical/biomedical transformers, each evaluated under three strategies: no fine-tuning, last-layer fine-tuning, and full fine-tuning. General-purpose models—particularly BERT and DistilBERT—improved substantially with last-layer fine-tuning, with BERT achieving the highest test F1 = 82.76 ± 4.51 on the ADReSSo Benchmark. By contrast, full fine-tuning often degraded performance (notably for RoBERTa and XLNet), consistent with overfitting on a limited dataset. Clinical/biomedical models showed smaller gains across all strategies, and although ClinicalBigBird and BlueBERT benefited modestly from full fine-tuning, their F1 scores remained below those of general-domain models. Overall, these findings suggest that pretraining on general-domain text better captures conversational disfluencies relevant to cognitive impairment than pretraining on structured clinical text.

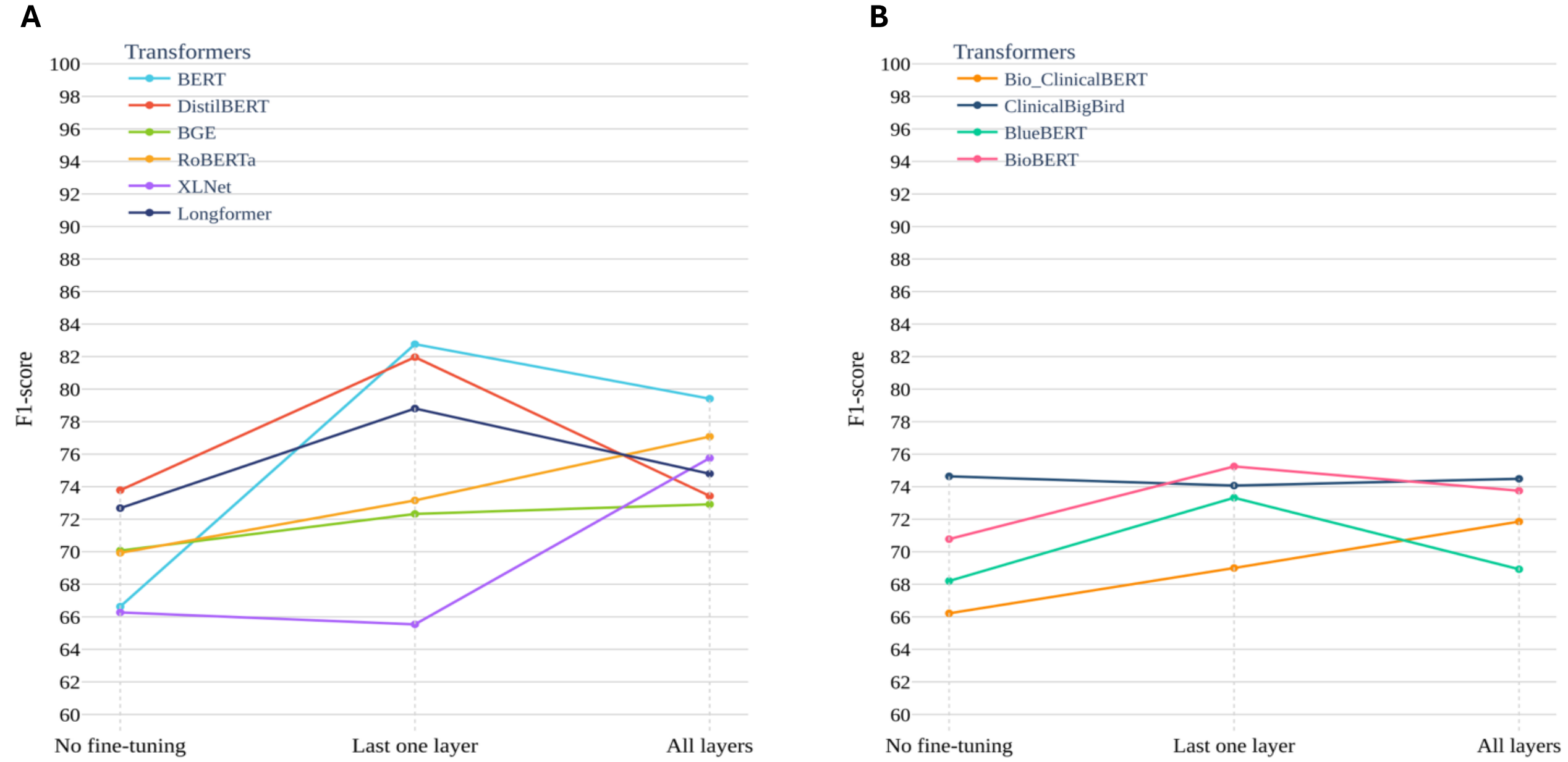

**Figure 3.** Performance of general-purpose and clinical-domain transformer models across fine-tuning strategies on the ADReSSo Benchmark. **A.** F1-scores of six general-purpose models across three strategies: no fine-tuning, last-layer fine-tuning, and full fine-tuning. BERT and DistilBERT showed the largest gains with last-layer fine-tuning. **B.** F1-scores of four clinical-domain models pretrained on biomedical and clinical corpora (e.g., PubMed, MIMIC-III). Performance gains were modest across all strategies, with overall F1-scores lower than those of general-purpose models.

### Performance of Handcrafted Linguistic Features

The classifier using 110 handcrafted linguistic features, capturing lexical richness/diversity, syntactic complexity, discourse fluency, and psycholinguistic categories, performed well on the validation set (F1 = 81.29) but did not generalize to the held-out test set (F1 = 66.83), indicating limited robustness to unseen speakers (Table 2).

### Performance of Late Fusion Classifier

Combining fine-tuned BERT embeddings with the same feature set in a late-fusion architecture reduced the generalization gap, achieving F1 = 83.32 and AUC = 89.48 on the test set (validation: F1 = 78.17, AUC = 82.05). Relative to the linguistic-only model, fusion improved test F1 by 24.7% and AUC by 19.6% (Table 2). Compared with BERT alone, fusion yielded a 0.7% increase in F1 with a −0.6% change in AUC (Table 2)

**Table 2.** Performance Comparison of BERT, Linguistic Feature-Based, and Fusion Models on Validation and Test Sets of the ADReSSo Benchmark

| Models | Validation F1<br>Mean ± 95%CI | Validation AUC<br>Mean ± 95%CI | Test F1<br>Mean ± 95%CI | Test AUC<br>Mean ± 95%CI |
|---|---|---|---|---|
| BERT | 78.97 ± 1.84 | 80.55 ± 1.26 | 82.76 ± 4.51 | 90.03 ± 1.29 |
| Linguistic features | 81.29 ± 1.16 | 78.10 ± 1.41 | 66.83 ± 4.32 | 74.83 ± 1.48 |
| Fusion Model<br>(BERT + Linguistic features) | 78.17 ± 1.29 | 82.05 ± 2.52 | 83.32 ± 2.78 | 89.48 ± 4.40 |

### 3.2 Component 2: LLM-Generated Synthetic Text for Augmentation

#### Per-Model Augmentation

Figure 4.A shows validation F1 after retraining the fusion model with synthetic transcripts from each LLM on the ADReSSo Benchmark. MedAlpaca-7B achieved the highest score (81.02), surpassing the baseline fusion model (F1 = 78.17, Table 2). LLaMA-8B and GPT-4 followed, while Ministral-8B provided modest gains and LLaMA-70B performed lowest (77.21), indicating that increased model capacity did not yield more informative synthetic data. These findings suggest that clinically tuned models such as MedAlpaca-7B generate synthetic speech that more effectively reinforces class-specific linguistic patterns.

Figure 4.B and 4.C compare synthetic to real transcripts using semantic similarity (BERTScore) and lexical overlap (BLEU-1–4) for this benchmark. LLaMA-8B achieved the highest BERTScore (0.61), reflecting strong semantic alignment. GPT-4 and LLaMA-70B followed (≈0.58). Although LLaMA-70B attained the highest BLEU-1 (0.87) and BLEU-2 (0.64), this did not translate into improved classification (Figure 4.A). MedAlpaca-7B maintained a high BERTScore (0.59) and consistently outperformed GPT-4 on BLEU, indicating that capturing class-specific structure and semantics is more important than surface lexical overlap.

Figure 4.D visualizes embeddings via t-SNE. Synthetic samples from MedAlpaca-7B and LLaMA-8B are interspersed with train, validation, and test data, consistent with their strong F1 and alignment metrics on this dataset. In contrast, GPT-4 and LLaMA-70B form distinct clusters, mirroring their lower semantic similarity, while Ministral-8B shows diffuse but less specific overlap. These patterns align with the performance differences observed in

#### Scaling Augmentation

Figure 5 evaluates the effect of augmentation scale for MedAlpaca-7B (Figure 5.A) and GPT-4 (Figure 5.B) on the ADReSSo benchmark. For GPT-4, a modest 1× augmentation yielded a slight improvement in F1 (83.32 ± 2.78 → 84.14 ± 1.92), but performance declined at 2× (80.76 ± 5.16) and fluctuated between 80–82 thereafter, consistent with t-SNE evidence of drift away from the real-speech manifold. In contrast, MedAlpaca-7B improved at 1× (85.35 ± 1.96), peaked at 2× (85.65 ± 1.64), and then declined with larger volumes (3× = 81.87 ± 3.03; 4× = 80.45 ± 4.36). These findings suggest that augmentation is beneficial only while synthetic data remains distributionally aligned, with effective limits of approximately 2× for MedAlpaca-7B and 1× for GPT-4. (For detailed results of 1× to 5× augmentation for both LLMs, see Appendix D)

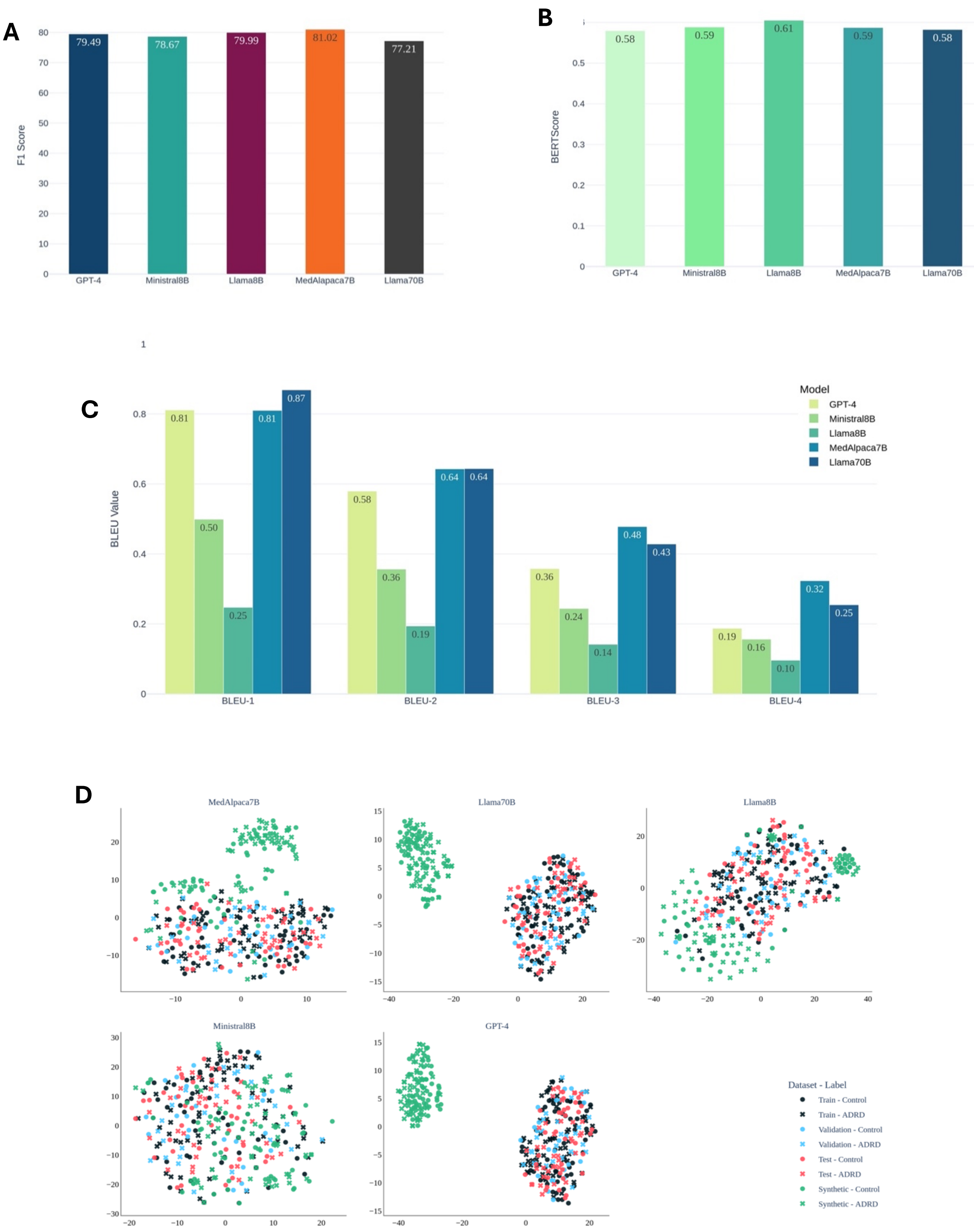

**Figure 4.** Evaluation of synthetic speech generated by LLMs for data augmentation on the ADReSSo benchmark. **A.** Validation F1-scores of the fusion-based screening model after augmenting the training set with LLM-generated transcripts; **B.** Semantic similarity of synthetic and human transcripts measured by BERTScore; **C.** Lexical similarity evaluated using BLEU-1 to BLEU-4 scores; **D.** visualization of the embedding space of original and synthetic

## A. MedAlpac

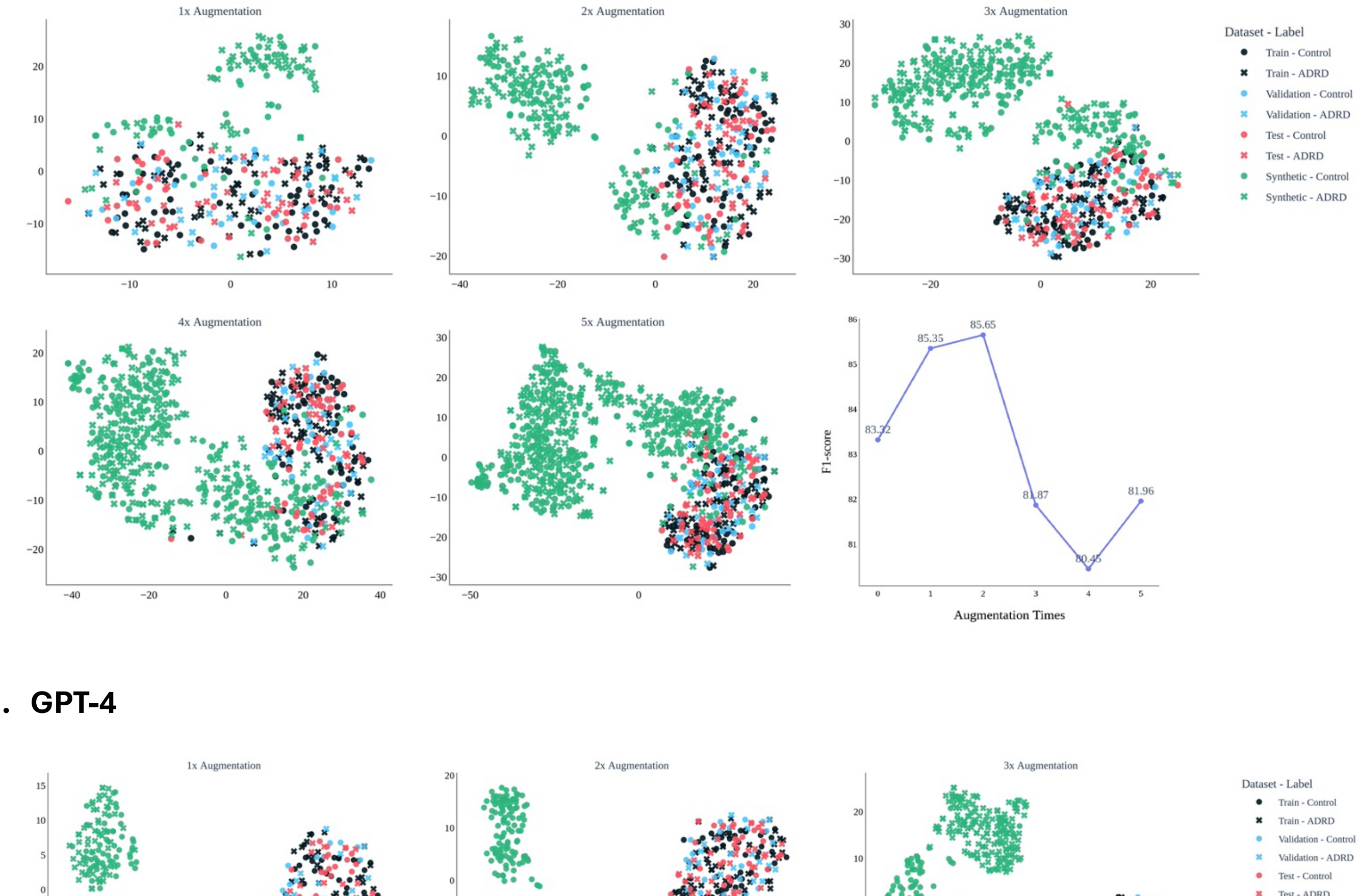

## B. GPT-4

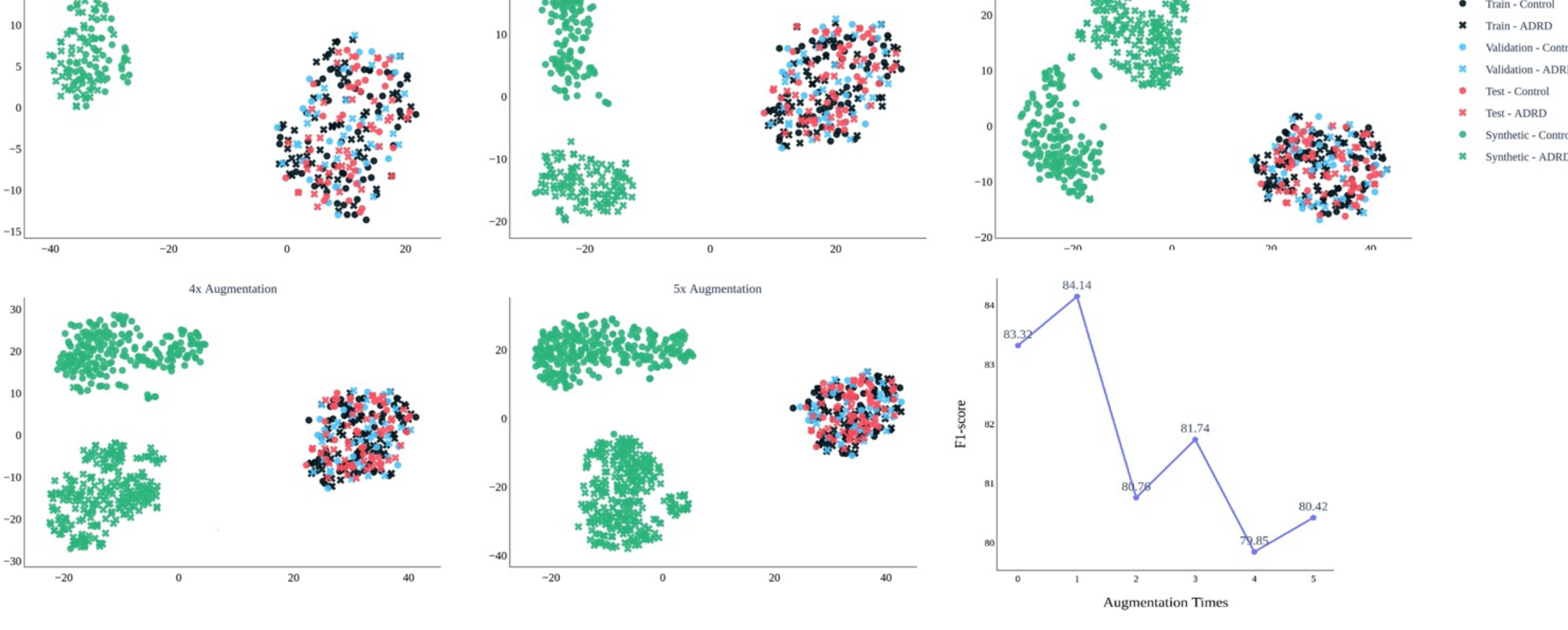

**Figure 5.** Effect of synthetic data volume on embedding structure and model performance. t-SNE plots show how synthetic narratives from MedAlpaca-7B (**5. A**) and GPT-4 (**5.B**) integrate with real data across 1× to 5× augmentation. MedAlpaca-7B remains aligned up to 2×, supporting peak F1 (85.7), while GPT-4 drifts after 1×, reducing effectiveness. Line plots show corresponding F1-scores of the fusion -based screening model on the held-out test dataset. The results are based on the ADReSSo benchmark.

### Operating Characteristics before vs after Augmentation

**Figure 6** compares the performance of the fusion model before (green) and after two-fold augmentation using MedAlpaca-7B synthetic speech (pink) on ADReSSo benchmark. The ROC curves (**Panel a**) remain nearly identical, with AUC improving marginally from 89.48 ± 4.40 to 89.56 ± 2.32, indicating stable overall discrimination. The precision–recall curve (**Panel b**) shows improved precision at lower recall levels, and the cumulative gains curve (**Panel c**) demonstrates enhanced early retrieval of positive cases, particularly between the 40%–70% sample range.

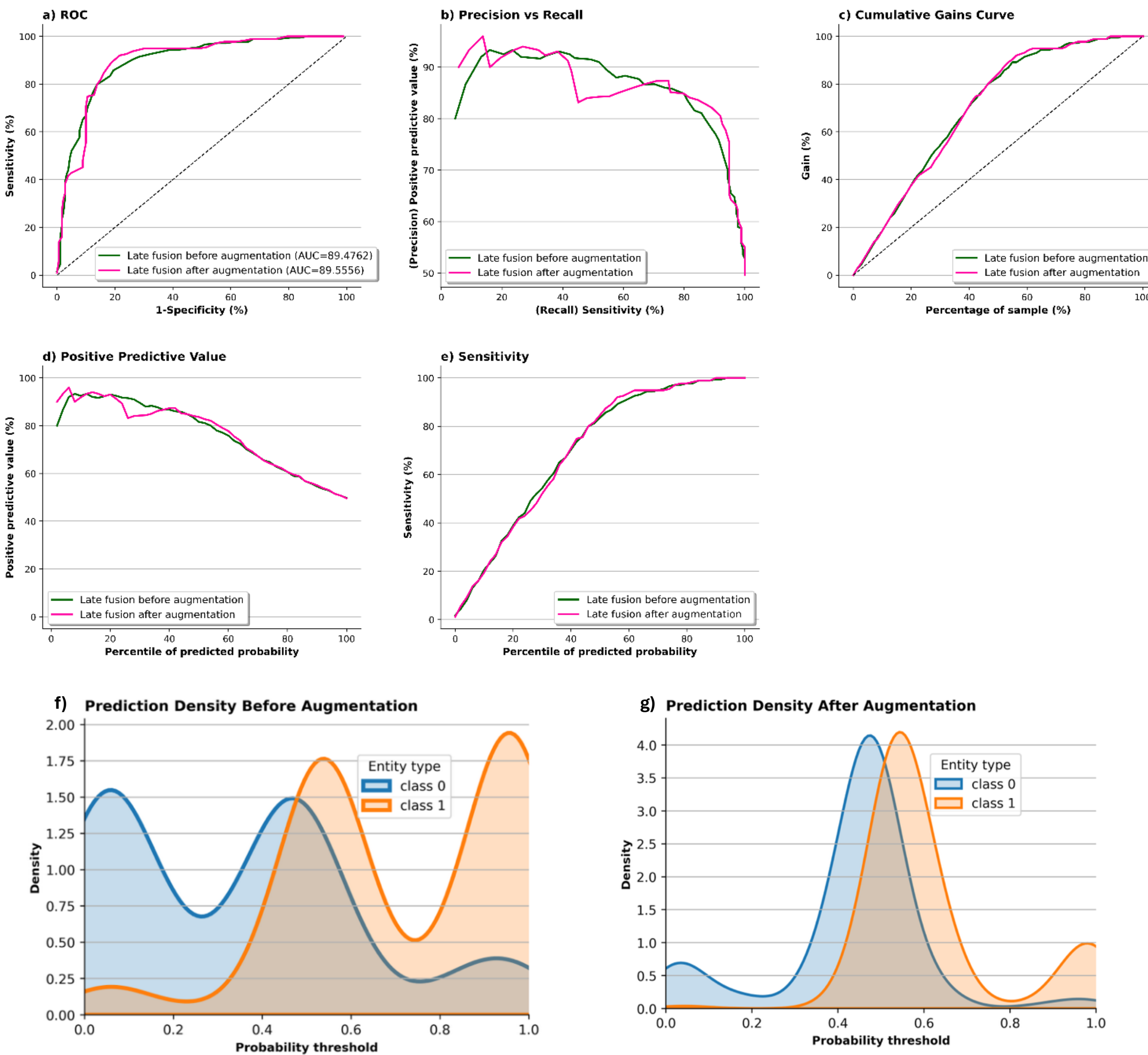

**Figure 6.** Impact of MedAlpaca-7B synthetic data on screening model performance and prediction confidence on ADReSSo benchmark. Panels a–e compare the fusion-based screening model before (green) and after 2× augmentation with MedAlpaca-7B synthetic speech (pink). ROC curves (**a**) show stable discrimination (AUC: 89.48 → 89.56). The precision–recall curve (**b**) shows improved precision at lower recall. Cumulative gains (**c**) indicate better early retrieval of positives (notably between 40%–70%). PPV (**d**) and sensitivity (**e**) profiles remain nearly identical. Density plots (**f–g**) show that post-augmentation predictions are more concentrated and better separated across classes, with reduced uncertainty around the 0.5 threshold.

**Panel d** indicates that the positive-predictive-value profiles of the pre- and post-augmentation models overlap across most probability percentiles, with only a slight dip for the augmented model at lower thresholds, while **Panel e** shows closely matching sensitivity curves, together implying that the added synthetic data left PPV largely intact and fully preserved sensitivity. The prediction density plots further support these findings: before augmentation (**Panel f**), class distributions overlapped considerably around the decision boundary, whereas after augmentation (**Panel g**), class 0 (cognitively healthy [control]) and class 1 (cognitively impaired [case]) predictions became more concentrated and more separable, with reduced uncertainty near the 0.5 threshold.

### 3.3 Components 3 and 4. Text-Based LLMs (Unimodal) and Multimodal (Audio +Text) LLMs Classifier

As shown in Table 3, which reports F1-scores with 95% confidence intervals, in the zero-shot setting, the best performance among text-only LLMs came from GPT-4o (F1-score = 73.05), followed closely by LLaMA 3.3 70B (72.93). For multimodal models, GPT-4o (omni) achieved 67.57, and Qwen 2.5 Omni reached 67.31.

With fine-tuning, all text-based LLMs improved. MedAlpaca-7B increased from 47.73 to 78.69 (improvement of 64.9%), and LLaMA 3.1 8B from 68.54 to 81.08 (+18.3%). More modest gains were observed for LLaMA 3.3 70B (+10.2%) and GPT-4o (+3.1%). Among multimodal models, Phi-4 showed one of the strongest improvements overall (+76.3%), whereas Qwen 2.5 Omni declined substantially (−26.8%).

Overall, Table 3 demonstrates that fine-tuning provides the greatest benefit for smaller or domain-specific text models, while multimodal models remain inconsistent, with outcomes ranging from major improvements (Phi-4) to sharp declines (Qwen 2.5 Omni).

**Table 3.** Zero-shot versus fine-tuned F1 performance of unimodal (text-only) and multimodal LLM classifiers on official test of ADReSSo benchmark

| Model | Zero-Shot F1 (%), Mean ± 95%CI | Fine-Tuned F1(%), Mean ± 95%CI | Improvement (%) |
|---|---|---|---|
| **Unimodal (text)** | | | |
| LLaMA 3.1 8B Instruct | 68.54 ± 2.1 | 81.08 ± 0.94 | + 18.3 |
| MedAlpaca 7B | 47.73 ± 1.03 | 78.69 ± 2.31 | + 64.9 |
| Ministral 8B (2410) | 66.58 ± 8.55 | 73.7 ± 3.08 | + 10.7 |
| LLaMA 3.3 70B Instruct | 72.93 ± 1.2 | 80.35 ± 1.92 | + 10.2 |
| GPT-4o (2024-08-06) | 73.05 ± 2.04 | 75.28 ± 0.95 | + 3.1 |
| **Multimodal** | | | |
| GPT-4o (omni) | 67.57 ± 1.08 | - | - |
| Qwen 2.5-Omni | 67.31 ± 0 | 49.3 ± 0 | - 26.8 |
| Phi-4 | 40.6 ± 3.77 | 71.59 ± 0.83 | + 76.3 |

### 3.4. Component 5. External Generalizability Evaluation — DementiaBank Delaware Corpus

**Fusion of Transformers and Handcrafted Linguistic Features.** Among the four transformers fine-tuned on the Delaware dataset, DistilBERT achieved the best performance (F1 = 66.02 ± 1.23). Integrating DistilBERT embeddings with handcrafted features in a late-fusion classifier further improved results, yielding F1 = 68.05 ± 3.16 and AUC = 62.90 ± 10.99.

**LLM-Based Synthetic Augmentation**. Incorporating MedAlpaca-7B synthetic samples with the original data improved performance with 1× augmentation (F1 = 72.82 ± 6.72; AUC = 69.57 ± 10.02). However, performance declined with 2× augmentation (F1 = 70.43 ± 2.27). These findings suggest that augmentation is beneficial only when synthetic data remains aligned with the real speech distribution, with effectiveness extending up to approximately 1× for MedAlpaca-7B on the Delaware corpus.

**Text-Based and Multimodal LLM Classifiers.** As shown in Table 4, among the text-only LLMs, GPT-4o achieved the highest improvement (+11.1%), while LLaMA 3.3–70B and LLaMA 3.1–8B showed smaller gains. For multimodal models, Phi-4 benefited most from fine-tuning (+17.6%). In contrast, GPT-4o (omni) was only evaluated in the zero-shot setting, as fine-tuning multimodal models was not possible through the API.

**Table 4.** Zero-shot versus fine-tuned F1 performance of unimodal (text-only) and multimodal LLM classifiers on held-out test set of Delaware

| Model | Zero-Shot F1 (%), Mean ± 95%CI | Fine-Tuned F1(%), Mean ± 95%CI | Improvement (%) |
|---|---|---|---|
| **Unimodal (text)** | | | |
| LLaMA 3.1 8B Instruct | 66.44 ± 1.54 | 68.23 ± 0.43 | + 2.7 |
| LLaMA 3.3 70B Instruct | 61.7 ± 2.21 | 64.52 ± 0.65 | + 4.6 |
| GPT-4o (2024-08-06) | 60.47 ± 2.82 | 67.18 ± 1.4 | + 11.1 |
| **Multimodal** | | | |
| GPT-4o (omni) | 60.23 ± 1.96 | - | - |
| Phi-4 | 55.76 ± 3.37 | 65.57 ± 0 | + 17.6 |

## 4. DISCUSSION

This study systematically evaluated pretrained transformer models and handcrafted linguistic features to develop a fusion model for early detection of cognitive impairment using spontaneous speech from the ADReSSo 2021 benchmark dataset. Prior studies have explored transformers and linguistic features separately, but our work is among the first to comprehensively assess ten transformer architectures across fine-tuning strategies and integrate their embeddings with 110 domain-informed linguistic features in a late-fusion design. Among the models tested, BERT with last-layer fine-tuning achieved the highest test performance (F1 = 82.76). Linguistic features alone showed strong internal validity (validation F1 = 81.29) but limited generalization (test F1 = 66.83), whereas the fusion of transformer embeddings and linguistic features improved robustness and achieved the strongest overall generalization (test F1 = 83.32).

Notably, BERT, one of the earliest transformer architectures, outperformed more recent and larger models. This result likely reflects both pretraining domain and dataset scale. BERT was trained on broad-domain English text that closely matches the conversational style of DementiaBank, while newer domain-specific models (e.g., BioBERT, ClinicalBERT) were optimized on biomedical documentation, which differs from spontaneous speech and is less effective at capturing disfluencies and syntactic irregularities. In addition, BERT's smaller architecture was better suited to the limited dataset size, whereas larger LLMs such as LLaMA-70B or GPT-4o typically require far larger task-specific corpora to realize their advantages and risk overfitting when fine-tuned on small samples. These findings suggest that for speech-based dementia detection, alignment between pretraining data, model capacity, and dataset scale may be more important than architectural novelty or size.

Building on this foundation, we further evaluated five state-of-the-art LLMs—LLaMA-3.3 8B, MedAlpaca, LLaMA-70B, Ministral, and GPT-4—and three leading multimodal models (text + speech) in both zero-shot and fine-tuned configurations. Zero-shot prompting allowed models to leverage latent knowledge of linguistic cues, whereas fine-tuning enabled adaptation to task-specific patterns. Substantial improvements followed fine-tuning, with the largest gains observed for MedAlpaca-7B (F1: 47.73 → 78.69), followed by LLaMA-8B and Ministral-8B, indicating that smaller open-weight models respond particularly well to targeted training. By contrast, multimodal LLMs underperformed, with Phi achieving the highest F1 = 71.59, suggesting that current audio–text architectures are not yet optimized for detecting cognitive-linguistic markers in spontaneous speech.

Training augmentation with LLM-generated transcripts was effective only when synthetic speech remained semantically and structurally aligned with real data. MedAlpaca-7B produced synthetic samples embedded within the real-data manifold (t-SNE overlap; BERTScore = 0.59), raising the fusion model's validation F1 from 78.17 to 81.02. The highest test F1 = 85.65 was achieved when synthetic data equaled twice the original training size; performance declined with further augmentation (3×–5×). LLaMA-8B and GPT-4 yielded smaller but stable improvements. In contrast, LLaMA-70B, despite strong lexical overlap (top BLEU scores), formed a separate embedding cluster and reduced performance to 77.2. These results confirm that lexical similarity alone is insufficient; effective augmentation must preserve cognitively salient features—such as repetition, disfluency, and syntactic errors—that carry diagnostic value. Thus, augmentation should be limited to volumes that maintain distributional alignment, accompanied by embedding-space validation to avoid degrading signal quality.

External evaluation on the Delaware corpus, restricted to clinically diagnosed MCI vs controls, provides strong evidence for early-stage screening. The late-fusion pipeline outperformed transformer-only and feature-only baselines, and limited augmentation with MedAlpaca-7B (1×) further improved performance, whereas larger augmentation (≥2×) introduced distributional drift and reduced gains. These findings, combined with stable ROC characteristics, indicate that augmentation is effective only within bounded scales and should be accompanied by embedding-space monitoring. Importantly, successful transfer to an MCI-only cohort demonstrates the pipeline's generalizability beyond the mixed-severity ADReSSo benchmark, directly addressing concerns about disease severity and reinforcing its relevance for early detection. Future work should expand MCI samples across sites and incorporate standardized prompts with pre-specified calibration to improve transportability.

Recent regulatory advances underscore the growing relevance of multimodal screening. In May 2025, the FDA approved Fujirebio's Lumipulse G pTau217/β-amyloid 1-42[52] blood test for Alzheimer's disease, offering a minimally invasive biomarker assay. While biologically informative, such tests do not reflect how cognitive decline manifests in everyday communication. Language changes—reduced fluency, disorganized sentences—often appear early and may signal real-world functional decline that biological tests cannot detect. Transformer models and LLMs offer a scalable solution by analyzing short voice recordings to identify subtle communication deficits. Their ability to detect linguistic cues offers a critical complement to biomarker testing.[53,54] Combining biological data with speech-based analysis may yield a fuller clinical picture, supporting earlier and more informed decisions on referrals, imaging, and intervention.

The potential of speech processing algorithms for cognitive screening in healthcare is significant, emphasizing the need for comprehensive research on its integration into clinical workflow.[55] This calls for interdisciplinary studies to understand clinical facilitators and barriers, including compatibility with existing workflows, clinician attitudes, and operational challenges. It is essential to evaluate the technical, logistical, and financial viability of deploying speech-processing tools in clinical settings, considering their fit with current practices and their ability to improve cognitive health assessments.[53,54] Overcoming these challenges by leveraging government support is essential for harnessing AI's potential to advance patient care and outcomes for patients with cognitive impairment.[54]

Some ASR systems normalize transcripts and suppress diagnostic cues.[13,56] In our prior benchmarking on the Pitt audio, Amazon Transcribe (general) maintained a competitive English WER (~13%) and preserved verbatim cues (fillers, repetitions, fragmented/partial words) more reliably than other transcription systems (Amazon Whisper, wav2vec2). For real-world automatic screening pipeline, we recommend ASR settings that retain disfluencies/partial words and avoid grammar-rewriting post-processors, with periodic audits of cue rates and threshold recalibration as ASR systems evolve.

Research on digital linguistic biomarkers spans a continuum from handcrafted linguistic features to transformer-based text models, acoustic and multimodal approaches, and, more recently, LLM-centric methods.[57,58] On the DementiaBank "Cookie Theft" task, early transcript-only studies relied on feature sets indexing lexical richness/diversity, syntactic complexity, discourse coherence, and filler/repetition rates, establishing that language organization itself provides clinically informative signal.[15 59] Building on this foundation, BERT and its derivatives applied to transcripts—typically with no tuning or last-layer fine-tuning—achieved performance in the range of ≈75–84 (F1/accuracy), capturing disfluencies and local syntactic irregularities more effectively than purely hand-engineered sets.[60] Other transcript-based approaches explored within the same benchmark include stacking ensembles of linguistic complexity, disfluency features, and pretrained transformers[61] (F1 ≈ 82), as well as functionals of deep textual embeddings enriched with silence-segment preprocessing and fusion model[62] (F1 ≈ 84.45). In parallel, acoustic and multimodal methods have combined prosodic or MFCC-based features with representations from speech transformers such as wav2vec 2.0 or Whisper, often using late or attention-based fusion with text embeddings. These systems commonly report performance in the ≈82–87 range, depending on the linguistic model, acoustic feature extraction, and fusion strategy.[63–65] More recently, LLMs have been explored in zero-shot, few-shot, or fine-tuned configurations for tasks such as dementia detection or disfluency scoring, although their clinical utility remains under investigation.[66] Within this landscape, our BERT (last-layer fine-tuned) achieved F1 = 82.76, matching or exceeding state-of-the-art LLM fine-tuning baselines; late fusion with handcrafted linguistic features increased performance to F1 = 83.32. With distribution-aligned synthetic augmentation, the fusion model further improved to F1 = 85.65, surpassing all LLM and audio-LLM fine-tuning baselines in our evaluation.

**Limitations**

This study has several limitations. First, we focused primarily on textual representations of speech and did not include acoustic transformer models, such as Whisper or wav2vec 2.0, which may capture additional prosodic or phonatory cues relevant to cognitive impairment. Second, while we evaluated several leading unimodal and multimodal LLMs, the multimodal models were limited to those with open or partially accessible APIs, and we could not fully fine-tune closed models like Google Gemini. Third, our evaluation was restricted to English-language transcripts from a structured task, which may limit generalizability to more diverse or naturalistic speech settings. Finally, although we used standard metrics and t-SNE for embedding analysis, future work should incorporate more rigorous interpretability methods to examine which linguistic and acoustic features drive model predictions.

## 5. CONCLUSION

This study demonstrates the potential of combining transformer-based embeddings with handcrafted linguistic features to improve the early detection of cognitive impairment from spontaneous speech. By systematically evaluating a range of pretrained transformer models and LLMs—including both unimodal and multimodal architectures—we identified configurations that maximize generalization and classification performance. Our results show that clinically tuned LLMs like MedAlpaca-7B not only adapt well to task-specific fine-tuning but also generate synthetic speech that meaningfully augments training data when distributionally aligned with real speech. These findings support the use of speech-based AI tools as a scalable and interpretable complement to biomarker-driven approaches for dementia screening and highlight the need for continued development of linguistically sensitive, clinically integrated NLP models.

Future work will build on these findings by addressing current limitations. We plan to integrate acoustic transformer models (e.g., wav2vec 2.0, Whisper) to capture prosodic and phonatory cues, enhance multimodal development through open-weight models with distillation from closed systems, and extend evaluation to multilingual and naturalistic conversations through multi-site validation. We will also strengthen the augmentation pipeline by combining LLM-guided narrative generation with prosody-controllable text-to-speech (TTS) to produce distributionally aligned synthetic speech. In addition, we aim to improve interpretability with feature attribution, psycholinguistic mapping, and fairness audits, providing insights into model decisions. Together, these efforts will help overcome current constraints and move this line of work toward practical, clinically useful screening tools.

**Author Contributions:**

- Ali Zolnour: Data processing; development of the fusion model; data augmentation; drafting the manuscript
- Hossein Azadmaleki: Development of the fusion model; drafting the manuscript
- Yasaman Haghbin: Development of the fusion model; data augmentation using LLMs; drafting the manuscript
- Fatemeh Taherinezhad: Evaluation of unimodal and multimodal LLMs; drafting the manuscript
- Mohamad Javad Momeni Nezhad: Data augmentation using LLMs; evaluation of unimodal and multimodal LLMs
- Sina Rashidi: Evaluation of multimodal LLMs
- Masoud Khani: Data augmentation using LLMs
- Amir Sajjad Taleban: Data augmentation using LLMs
- Samin Mahdizadeh Sani: Contributed to code documentation; manuscript review
- Maryam Dadkhah: Contributed to figure design; manuscript review
- James M. Noble: Manuscript review
- Suzanne Bakken: Manuscript review
- Yadollah Yaghoobzadeh: Contributed to methodology development; manuscript review
- Abdol-Hossein Vahabie: Contributed to methodology development; manuscript review
- Masoud Rouhizadeh: Manuscript review
- Maryam Zolnoori: Led conceptual model design; drafted and critically revised the manuscript

## Declarations

**Ethics Approval declaration:** Both datasets used in this study were obtained from DementiaBank, a publicly available resource hosted by TalkBank (https://dementia.talkbank.org/). The original data collection was approved by the Institutional Review Board of the University of Pittsburgh. As this study involved secondary analysis of de-identified data, no additional IRB approval was required.

**Human Ethics and Consent to Participate declaration:** Human Ethics and Consent to Participate declarations: not applicable.

## Appendix A. Reported performance of transformer models on the DementiaBank dataset across different studies

Transformer-based architectures, such as BERT, employ self-attention to generate contextualized word embeddings that capture both semantic and syntactic information. Trained on large-scale corpora, BERT and its derivatives have advanced clinical NLP by modeling complex linguistic patterns. On DementiaBank's "picture-description task," these models have shown strong performance (F1: 75–87.5), although outcomes differ based on fine-tuning approaches (e.g., updating only the last layer vs. the entire model), classifier types (binary vs. multi-layer perceptron), and use of validation data. Some studies found general-purpose BERT outperforming domain-specific transformers, highlighting a lack of systematic evaluation regarding which configurations best capture linguistic cues of cognitive impairment.

**Table 4. Reported performance of transformer models on the DementiaBank dataset across different studies**

| Study | Key Findings |
| --- | --- |
| Pappagari et al. [67] | Utilized Gradient Boosting machines with BERT embeddings, achieving 75% accuracy. |
| Koo et al.'s [68] | Trained a CCN Network on the word embeddings obtained from XLNet transformer model, leading to an 81.25% accuracy. |
| Balagopalan et al. [60] | Employed SVM classifiers on BERT, achieving an accuracy of 81.8%. |
| Zhu et al. [69] | Fine-tuned various BERT models (base and large), with the Longformer model reaching the highest accuracy at 82.08% using a multilayer perceptron. |
| TaghiBeyglou et al. [70] | Fine-tuned several models, including BERT and BioClinicalBERT, achieving the highest performance for BioClinicalBERT with an accuracy of 84% utilizing a multilayer perceptron. |
| Ilias et al. [71] | Fine-tuned several transformer models, including BERT, BioBERT, BioClinicalBERT, ConvBERT, RoBERTa, ALBERT, and XLNet. Finding BERT to yield the highest classification performance at 87.50% accuracy. |

## Appendix B. Prompt Template for Zero-Shot and Fine-Tuned LLM-Based Classification of Speech Transcripts as Cognitively Healthy or Impaired (ADRD)

*You are an expert in cognitive health and language analysis. You will analyze a spoken language transcript from a person describing the 'cookie theft' picture. This is not written text but a transcription of spontaneous speech. Analyze the provided transcript and classify it into one of two categories: 'Healthy' for a healthy cognitive state or 'AD' for Alzheimer's disease. Provide only the label ('Healthy' or 'AD') as the output. Do not include explanations or additional text. The output should be in JSON format, like {'label': 'predicted label'}.*

**Table 5. The optimal configuration for each LLM**

| Model | QLoRA Rank | QLoRA Alpha | QLoRA Dropout | Effective Batch Size | Epochs |
|---|---|---|---|---|---|
| LLaMA 8B | 64 | 128 | 0.1 | 8 | 12 |
| MedAlpaca 7B | 128 | 256 | 0.1 | 8 | 6 |
| Ministral 8B | 32 | 64 | 0 | 8 | 10 |
| LLaMA 70B | 16 | 32 | 0 | 8 | 9 |
| GPT-4o | - | - | - | 20 | 10 |

## Appendix C. Prompt Ablation: Effect of Label-Specific Cues in Fine-Tuning and Inference

We conducted a prompt ablation study to evaluate the effect of label-specific linguistic cues on model performance. Three configurations were compared:

1. **No cues in fine-tuning or inference:** The model trained without cues at either stage achieved F1 = 77.80 ± 4.15 and AUC = 85.90 ± 2.23, reflecting a notable decline relative to our primary method.
2. **Cues in both fine-tuning and inference:** When cues were incorporated into both stages, performance declined further (F1 = 77.59 ± 4.65, AUC = 82.94 ± 4.59).
3. **Cues in fine-tuning only (our method):** Our strategy—introducing cues during fine-tuning but omitting them during inference—yielded the strongest and most generalizable results (F1 = 85.65 ± 1.64, AUC = 89.56 ± 2.32).

These findings indicate that label-specific cues are most effective for enhancing representation learning during fine-tuning. However, their presence in inference prompts reduces naturalness and generalizability. Thus, the optimal approach is to include cues during training but exclude them during inference.

## Appendix D. Control Augmentation Experiments: Random Deletion ("Mask Augment") versus LLM-Based Augmentation

To evaluate whether the observed performance gains could be attributed to spurious augmentation effects, we conducted control experiments using a simple non-LLM perturbation method. Specifically, we applied random deletion ("Mask augment"), in which words were randomly removed from transcripts to generate synthetic data. We compared this approach with distribution-aligned LLM-based augmentation (GPT-4, MedAlpaca-7B) across augmentation scales ranging from 1× to 5× (i.e., synthetic data equal to one to five times the size of the training set).

As summarized in Table 6, Mask augment provided no benefit across scales. Its performance fluctuated below the baseline (83.32 ± 2.78) and declined at larger augmentation factors, showing no consistent improvements. By contrast, LLM-based augmentation yielded stable gains at small scales (1×–2×), reflecting its ability to generate linguistically coherent and distribution-aligned data.

These findings confirm that the improvements in our method are not due to arbitrary perturbations but instead arise from the ability of LLM-generated narratives to capture realistic linguistic structures and distributional patterns relevant to cognitive status classification.

**Table 6. Comparison of control augmentation (random deletion, "Mask augment") and LLM-based augmentation (GPT-4, MedAlpaca-7B) across augmentation scales (1×–5×).**

| Model | Data Augmentation F1 (Mean + 95%CI) | | | | |
|---|---|---|---|---|---|
| | One | Two | Three | Four | Five |
| **Mask augment** | 79.47 ± 4.06 | 81.39 ± 2.25 | 80.42 ± 2.25 | 78.81 ± 1.44 | 78.12 ± 4.62 |
| **GPT-4** | 84.14 ± 1.92 | 80.76 ± 5.16 | 81.74 ± 3.35 | 79.85 ± 2.69 | 80.42 ± 5.17 |
| **MedAlpaca** | 85.35 ± 1.96 | 85.65 ± 1.64 | 81.87 ± 3.03 | 80.45 ± 4.36 | 81.96 ± 2.77 |